\documentclass[conference]{IEEEtran}

\IEEEoverridecommandlockouts
\usepackage{cite}
\usepackage{amsmath,amssymb,amsfonts}
\usepackage{graphicx}
\usepackage{textcomp}
\usepackage{xcolor}
\def\BibTeX{{\rm B\kern-.05em{\sc i\kern-.025em b}\kern-.08em
    T\kern-.1667em\lower.7ex\hbox{E}\kern-.125emX}}

\usepackage{algorithm}
\usepackage[noend]{algorithmic}
\usepackage[utf8]{inputenc}
\usepackage{amsmath}
\usepackage{amsfonts}
\usepackage{mathtools}
\usepackage{booktabs} 
\usepackage{xcolor}
\usepackage{comment}
\usepackage{amsmath,graphicx}
\usepackage{multirow}
\usepackage{amssymb}
\usepackage{mathtools,amsthm}
\usepackage{url}
\usepackage{epstopdf}
\usepackage{amsthm}
\usepackage{enumitem}
\usepackage{lipsum}
\usepackage{soul}
\usepackage{subfig}
\usepackage{times}
\usepackage{makecell}
\usepackage{color,soul}
\soulregister\cite7
\soulregister\ref7
\soulregister\pageref7
\usepackage{diagbox}
\usepackage[export]{adjustbox}
\renewcommand{\footnotesize}{\scriptsize}
\usepackage{bm}

\usepackage{soul}
\usepackage{comment}
\usepackage{caption}
\usepackage{lipsum}
\usepackage{microtype}
\newcommand\eat[1]{}
\usepackage{footnote}
\makesavenoteenv{tabular}
\usepackage{setspace}
\usepackage{titlesec}
\usepackage{amsthm}

\newtheorem{assumption}{Assumption}
\newcommand{\cmt}[1]{{\color{black} \small \# #1 }}

\makeatletter 
\newcommand{\linebreakand}{%
  \end{@IEEEauthorhalign}
  \hfill\mbox{}\par
  \mbox{}\hfill\begin{@IEEEauthorhalign}
}
\makeatother 

\newtheorem{innercustomthm}{Theorem}
\newenvironment{customthm}[1]
  {\renewcommand\theinnercustomthm{#1}\innercustomthm}
  {\endinnercustomthm}

\usepackage[left=1.62cm,right=1.62cm,top=1.8cm]{geometry}

\newcommand\x[1]{\textcolor{black}{#1}}

\begin{document}

\title{FedX: Adaptive Model Decomposition and Quantization for IoT Federated Learning
}


\author{\IEEEauthorblockN{*Phung Lai}
\IEEEauthorblockA{
\textit{University at Albany}\\
Albany, New York, USA \\
lai@albany.edu}
\and
\IEEEauthorblockN{*Xiaopeng Jiang}
\IEEEauthorblockA{
\textit{Southern Illinois University}\\
Carbondale, Illinois, USA \\
xiaopeng.jiang@siu.edu}
\and
\IEEEauthorblockN{ NhatHai Phan, Cristian Borcea, Khang Tran}
\IEEEauthorblockA{
\textit{New Jersey Institute of Technology}\\
Newark, New Jersey, USA \\
phan@njit.edu, borcea@njit.edu, kt36@njit.edu}
\and
\linebreakand
\IEEEauthorblockN{An Chen, Vijaya Datta Mayyuri}
\IEEEauthorblockA{
\textit{Qualcomm Incorporated}\\
San Diego, California, USA \\
anc@qualcomm.com,vmayyuri@qualcomm.com}
\and
\IEEEauthorblockN{ Ruoming Jin}
\IEEEauthorblockA{
\textit{Kent State University} \\
 Kent, Ohio, USA \\
rjin1@kent.edu}
}

\maketitle
\begingroup\renewcommand\thefootnote{\textsection}
\footnotetext{* Equal contribution}
\endgroup

\begin{abstract}
Federated Learning (FL) allows collaborative training among multiple devices without data sharing, thus enabling privacy-sensitive applications on mobile or Internet of Things (IoT) devices, such as mobile health and asset tracking. However, designing an FL system with good model utility that works with low computation/communication overhead on heterogeneous, resource-constrained mobile/IoT devices is challenging. 
To address this problem, this paper proposes \textbf{FedX}, a novel adaptive model decomposition and quantization FL system for IoT. To balance utility with resource constraints on  IoT devices, FedX decomposes a global FL model into different sub-networks with adaptive numbers of quantized bits for different devices. The key idea  is that a device with fewer resources receives a smaller sub-network for lower overhead but utilizes a larger number of quantized bits for higher model utility, and vice versa. The quantization operations in FedX are done at the server to reduce the computational load on devices. FedX iteratively minimizes the losses in the devices' local data and in the server's public data using quantized sub-networks under a regularization term, and thus it maximizes the benefits of combining FL with model quantization through knowledge sharing among the server and devices in a cost-effective training process. Extensive experiments show that FedX significantly improves quantization times by up to 8.43$\times$, on-device computation time by 1.5$\times$, and total end-to-end training time by 1.36$\times$, compared with baseline FL systems. We guarantee the global model convergence theoretically and validate local model convergence empirically, highlighting FedX's optimization efficiency. 
\end{abstract}


\section{Introduction}


In Federated Learning (FL), a global model is trained by aggregating local model's gradients across devices, while ensuring user privacy by not sharing local data \cite{bonawitz2019towards}. This feature makes FL suitable for building intelligent mobile/IoT systems and enhancing model generalization/utility, compared with individual local models. Examples include mobile health, where  users employ heterogeneous IoT devices  (i.e., phone, smart watches,  mHealth sensors) to monitor health information and human activities  \cite{wang2023applications}, and fleet tracking, where  businesses use different IoT devices to track assets or vehicles \cite{thorgeirsson2021probabilistic}. In these systems, the devices exhibit heterogeneity, but collaborate to achieve the  same system goals by performing identical tasks.

The hardware heterogeneity of IoT devices in terms of computation, communication, and storage raises a main challenge for FL systems: How to balance model utility with computation and communication overhead, given heterogeneous and limited on-device resources?
Achieving high model utility may require resource-intensive models and multi-round communication, straining resource-constrained devices.  
Large models can be too slow, difficult to execute, or consume too much battery power, while small models on all devices may overlook that some devices can handle the larger models that perform better. 

Model compression and distillation address computation and communication overhead on resource-constrained devices. Quantization-based compression approaches \cite{honig2022dadaquant,oh2023fedvqcs} reduce  communication load by reducing parameter or gradient precision.   
Other compression methods are sparsification, pruning,  reduction, and sampling \cite{hamer2020fedboost,jiang2023computation}. However, these  techniques often sacrifice  utility or focus on communication reduction at the expense of computation.
In model distillation \cite{li2019fedmd}, different devices with heterogeneous resources initially train their model using public data, then retrain locally using their private data. To mitigate the reliance on public data,  some works~\cite{li2021talk,liu2023communication} leverage quantization on devices and conduct model shrinking or knowledge distillation to align  utility  of the local model with that of the global model.
While effective, these techniques can impose significant computational overhead on clients.  In addition,  knowledge distillation, tiering, and split learning-based FL \cite{mohammadabadi2023speed,zhang2023privacy,li2019fedmd} focus on individual devices without globally balancing utility,   computation, and communication. Therefore, there is a crucial need for an efficient and effective FL for heterogeneous and resource-constrained IoT devices. 




\newpage

\textbf{Contributions.} This paper proposes \textbf{FedX}, an adaptive model decomposition and quantization FL system for IoT. To balance model utility with resource constraints on IoT devices, FedX decomposes a global FL model into different sub-networks 
with different quantization levels for different devices. 
The key idea to achieve this balance is that a device with fewer resources receives a smaller sub-network for lower overhead, but utilizes a larger number of quantized bits for higher model utility, and vice-versa.
The sub-networks used by the devices are nested within each other; that is, a smaller sub-network is always contained inside a larger sub-network. This approach fosters knowledge sharing among IoT devices, since on-device training and updating any sub-network will improve other sub-networks through the model aggregation in the federated training.
Quantization operations in FedX are done at the server to reduce the computation load on IoT devices, notably without affecting their model utility.

FedX starts from a global model trained on publicly available data and fine-tunes it to align it with the local data distributions of IoT devices, which frequently differ from the global model's distribution.
During this process, reducing the disparity between the global models before and after fine-tuning is crucial to ensure that the fine-tuned global model effectively learns the distinctive distribution present in the server's data while retaining the knowledge acquired from the data distributions of the IoT devices. To ensure minimal disparity between the global models before and after fine-tuning, we add a regularization  minimizing the difference between the global models during fine-tuning. 
The features of FedX maximize the benefits of combining FL with model quantization by enabling knowledge sharing among the server and the devices in a cost-effective training process.

Extensive experiments on image classification and human activity recognition show that FedX outperforms state-of-the-art adaptive quantization approaches for FL~\cite{honig2022dadaquant,liu2023communication} by $1.90\%-12.28\%$ in accuracy. In addition, 
 using NVidia Jetson Nano, Qualcomm QCS605 AI kit, and Raspberry Pi 5, FedX shows significant improvement in quantization time by  $8.43\times$, on-device computation time by $1.5\times$, and total end-to-end training time by $1.36\times$, compared with baseline FL systems.

\section{Background}

\subsection{Federated Learning}
Federated Learning (FL) is a multi-round protocol where a server sends a global model $\theta^t$ to a random subset of $M^t$ devices at each training round $t$. These devices train the model locally and send updates back. The server aggregates the updates using a function $\mathcal{G}: R^{|M^t|\times n} \rightarrow R^n$ ($n$ is the size of $\theta^t$), producing the aggregated model $\theta^{t+1} =\mathcal{G}(\{\theta_u^{t+1}\}_{u\in M^t})$. FedAvg is a common aggregation method in FL \cite{kairouz2021advances}.

 

\subsection{Quantization-based  Compression}

Quantization-based compression approaches in FL typically speeds up model convergence and minimizes communication
\cite{alistarh2017qsgd,oh2023fedvqcs,liu2023communication}.  
In FedX, we consider stochastic fixed-point quantization, i.e.,  QSGD \cite{alistarh2017qsgd,honig2022dadaquant}. The quantization function, denoted as $\mathbb{Q}(\cdot, q)$  with $q \ge 1$, represents quantized bits, corresponding to  the number of bits used for each value. Technically,  $q$ is associated with $2^q$ levels of quantization. 
Each value is quantized in a manner that maintains its expected value while minimizing variance. 
The quantizer $\mathbb{Q}(\theta, q)$ quantizes the vector $\theta$ element-wise,  returning the sign of $\theta$ and $\| \theta\|_2$ rounded to one of the endpoints  of its encompassing interval. 

QSGD quantizes $\theta$ in three steps:
\textit{1)} Quantize $\theta$ as $\mathbb{Q}(\frac{\theta}{ \|\theta\|_2 })$ into $2^q$ bins in $[0,1]$, storing signs and $\|\theta\|_2 $ separately;  \textit{2)} Encode the results with $0$ run-length encoding; and \textit{3)} Encode the results with Elias $\omega$ coding. 
The first step is a lossy transformation, where   a higher $q$ lowers the loss, while subsequent steps are lossless. These steps reduce device computation and communication bits, including distributing updated models   between the server and devices and sending parameter/gradient updates. Quantization further reduces device computation by using fewer bits to represent values. 



\section{FedX: An Adaptive Quantization FL System}


\textbf{FedX Settings and Objectives.} 
We consider a setting in which the server's collected data $D_S$ and devices' local data $\{D_u\}_{u \in [1, N]}$ are complementary to each other but may not be in the same or similar distribution.
For instance, in asset tracking, devices can leverage local data to recognize more specialized and diverse objects than the server's public data.
The server and all the devices $u$ jointly train an aggregated global model $\theta$ and the devices' local models $\{\theta_u\}_{u \in [1, N]}$, such that the global and local models can recognize objects from both the client data and the server data after federated training. 
This configuration is practical as the server can access public data, which can supplement, but differ from the on-device data collected from various devices.

In this setting, FedX aims to enhance FL for heterogeneous IoT devices with adaptive decomposition and quantization, addressing: (i) device-tailored models balancing utility with resource constraints and overhead;  (ii) adaptive quantization during training; (iii) low overhead  on-device training; and (iv) computation savings at inference.  This approach improves model utility and system effectiveness while reducing training, quantization, and inference overhead.
Key challenges are: \textbf{(1)} balancing model utility with computation/communication overhead from device heterogeneity, and \textbf{(2)} optimizing knowledge transfer to reduce performance disparity and improve utility.

To address these challenges, departing from existing FL systems, the key idea of FedX is to decompose the global model $\theta$ into smaller sub-networks $\theta_u \subseteq \theta$ (Fig.~\ref{fig:subnets}), each tailored to a device's resources. Devices with fewer resources receive smaller sub-networks for lower training overhead, while using more quantized bits for higher utility and efficient inference; and vice-versa.   These local sub-networks are aggregated to update the global model $\theta = \mathcal{G} ({\theta_u}_{u \in [1, N]})$.




The optimization function for global and local models is:
\begin{small}
\begin{align}
    \theta^*  &= \arg\min_{\theta} \big[  \sum_{u \in [1,N]} \mathcal{L} ( \mathbb{Q}(\theta_u, q_u), D_u) + \mathcal{L}(\theta, D_S)  \big] \nonumber \\
    &\text{s.t. } \forall u \in [1, N]: \theta_u \subseteq \theta \text{ and } \theta = \mathcal{G} (\{\theta_u\}_{u \in [1, N]}), \label{objectivefunction}
\end{align}
\end{small}

\noindent where $\mathcal{L}$ is a loss function and $\mathbb{Q}$ is a quantization function, such as QSGD, with the numbers of quantized bits $\{q_u\}_{u \in N}$ and the local model $\{\theta_u\}_{u \in N}$ tailored to the device $u$. 
The first term $\sum_{u \in [1,N]} \mathcal{L} ( \mathbb{Q}(\theta_u, q_u), D_u)$ is the loss across devices given quantized local models $\mathbb{Q}(\theta_u, q_u)$ on their local model $\theta_u$ and data $D_u$; the second term $\mathcal{L}(\theta, D_S)$ is the loss of the aggregated global model $\theta$ on the server's data $D_S$.


In practice, the number of device types (as a function of their resource capabilities) is expected to be small; thus, instead of individual devices, we can consider categories of devices (similar amount of resources) that have the same sub-network or   number of quantized bits without affecting the generality of our objectives. Also, we do not consider a weighted correlation between the two losses since the server and the devices can optimize them independently in our training protocol.

\subsection{Adaptive Model Decomposition and Quantization} 


In FedX, the server decomposes $\theta$ into  a suitable sub-network $\theta_u$ and identifies a suitable number of quantized bits to each device $u$ based on its resources. Given   $\theta$, the potential sub-networks can be exponentially large. Some may fail to perform well or hinder knowledge sharing among devices, especially if they occupy different parts of the global model.

\begin{figure}[t] 
\centering
\subfloat[Sub-networks $\theta_1$ and $\theta_2$]{\label{fig:subnets-a}\includegraphics[scale=0.035]{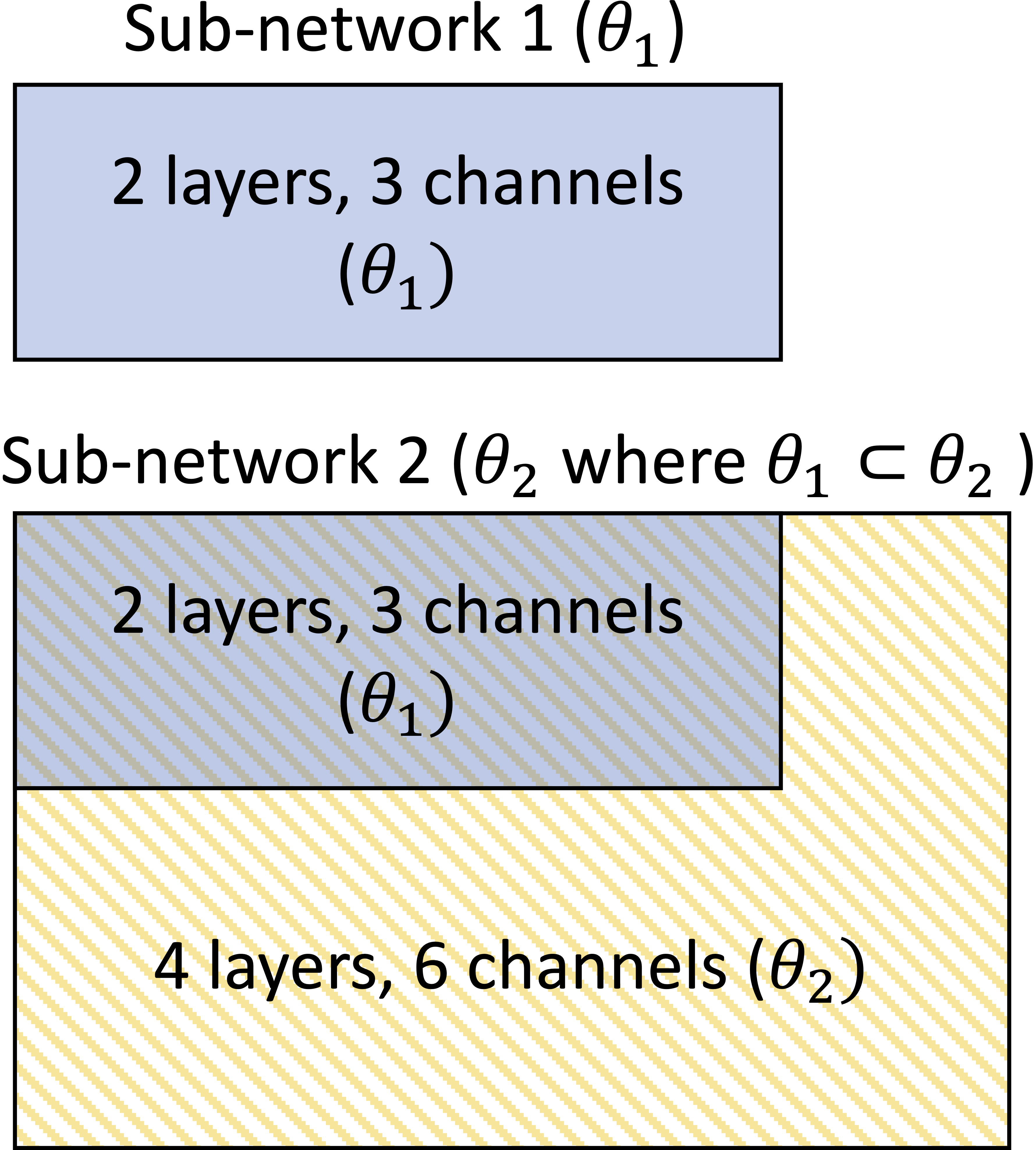}}\hfill
\subfloat[Aggregation of $\theta_1$ and $\theta_2$]{\label{fig:subnets-b}\includegraphics[scale=0.035]{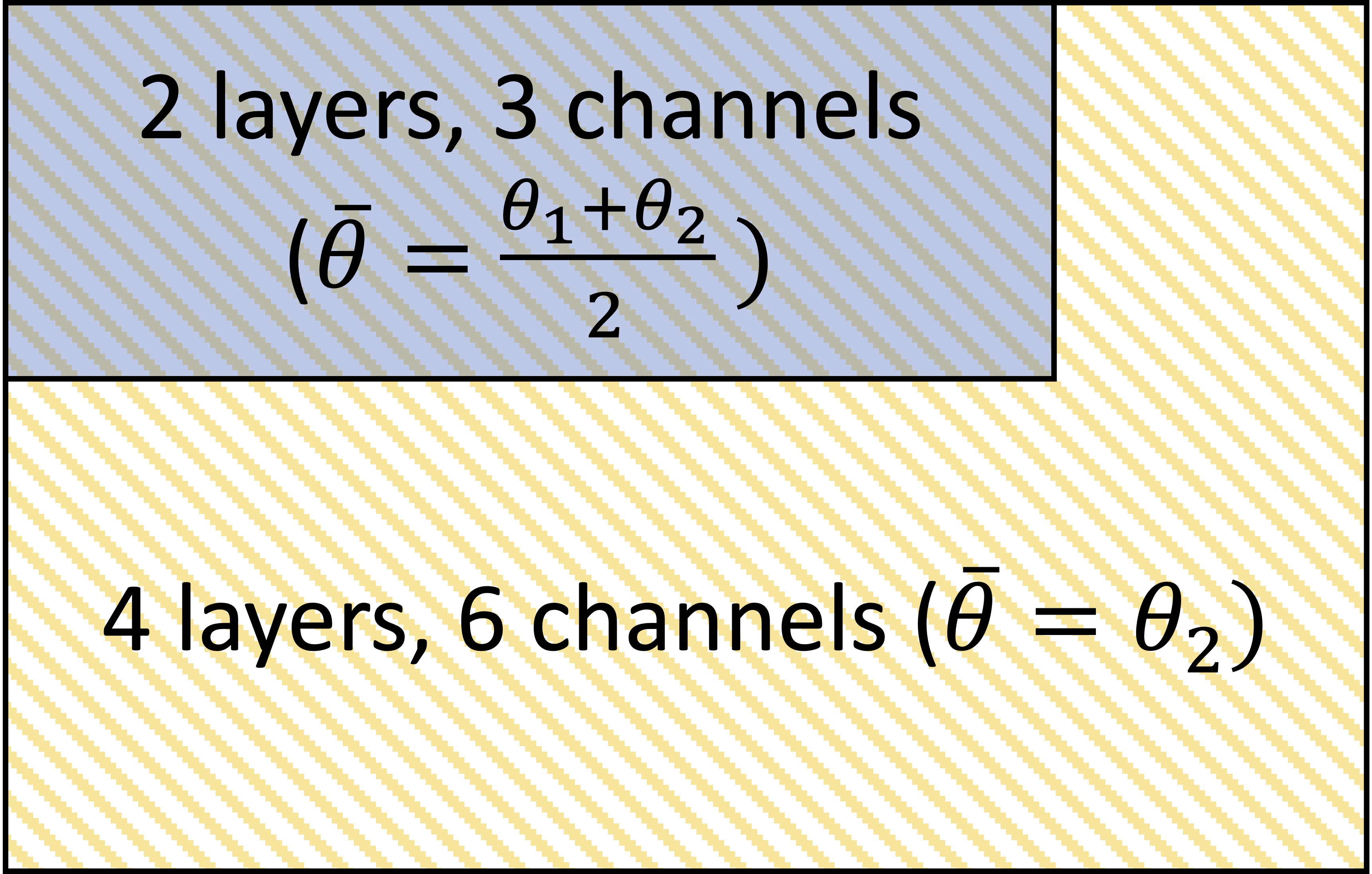}}\hfill 
\vspace{-5pt}
\caption{Nested sub-networks and their aggregation.
} 
\label{fig:subnets}  
\end{figure} 
\setlength{\textfloatsep}{5pt}



To address this problem, the server generates and trains a small set of sub-networks using its publicly available data $D_S$. 
In FedX, a smaller sub-network is always contained inside a larger sub-network. We call this setting \textit{\textbf{nested sub-networks}}. For instance, we consider the 
two sub-networks  $\theta_1$ and $\theta_2$ (Fig.~\ref{fig:subnets-a}). 
The sub-network $\theta_1$  has 2 layers, each of which has 3 channels. The sub-network $\theta_2$ has 4 layers, each of which has 6 channels. The sub-network $\theta_1$ is a subset of model parameters given the sub-network $\theta_2$ such as $\theta_1 \subset \theta_2$.


For instance, in our experiments with images,  we use CompOFA \cite{sahni2021compofa} with MobileNetV3 \cite{howard2019searching} as the global model $\theta$. CompOFA is a centralized training approach to generate and train sub-networks with varying sizes  from a larger model for efficient IoT deployment, reducing potential sub-networks while maintaining utility. The server generates 243 nested sub-networks ranging from 2.41M to 6.39M parameters, achieving good utility on server data $D_S$. However, due to differing data distributions between devices and the server, these sub-networks may not perform well on devices' data ${D_u}_{u \in [1, N]}$ before federated training. This nested sub-network approach generalizes beyond CompOFA and MobileNetV3, as demonstrated in our human activity recognition experiments.



Using a small set of nested sub-networks as local models $\theta_u$ enables effective knowledge sharing among IoT devices, as updates in any sub-network benefit others during federated training. Our observations indicate that failure  to construct these nested sub-networks significantly damages model utility, as non-nested updates from different clients can cause  misalignment among them. 
While neural architecture search can yield high-accuracy and efficient architectures \cite{jiang2019accuracy}, it requires repeating the search process and retraining for new hardware platforms, limiting scalability. In addition, individually trained models with no shared weights result in a large total model size that requires  high download bandwidth requirements \cite{cai2019once}.




To select a suitable sub-network and identify the  quantized bits for device $u$, we use the following  metrics and process:




\textbf{Selection Metrics.} \textit{1) Utility Drop:} This is the difference in prediction accuracy between the global model and a particular (either original or quantized) sub-network. \textit{2) On-device Training Time:} It is the training time per local round on a device using a sub-network. 

Although a sub-network can have efficient on-device training time, quantization\footnote{Note that quantization does not affect on-device training time in FedX since quantization will be conducted by the server, as discussed in the FedX training protocol.} may significantly degenerate its model utility. 
Hence, a suitable sub-network $\theta_u$ for a device $u$ must balance the trade-off between performance drop, on-device training time, and the number of quantized bits $q_u$. We formulate this trade-off as follows:
\begin{small}
\begin{align}
\theta_u, q_u &= \arg\min_{\theta_u \in \Theta, q_u \in [1, q^{max}_{u}]} \mathcal{P} (\theta, D_S) - \mathcal{P} (\mathbb{Q}(\theta_u, q_u), D_S) \nonumber \\
&\text{s.t. } \mathcal{T}(\theta_u, D_S) \leq \mu_u.
\label{selectionObjective}
\end{align}
\end{small}


\noindent where $\mathcal{P}(\theta, D_S)$ is the prediction accuracy of the model $\theta$ on the dataset $D_S$, $\Theta$ is the set of generated sub-networks, and $q^{max}_u$ is the maximum number of quantized bits that adhere to the processor's bit-width of  device $u$ . For instance, if device $u$  supports $16$ bits,  the maximum number of quantized bits $q^{max}_u$ is $16$. $\mathcal{T}(\theta_u, D_S)$ is the on-device training time of the sub-network $\theta_u$, and $\mu_u$ is a threshold, e.g., 10 seconds/local round, predefined by the server for efficient FL.
 It is worth noting that our  sub-network design prioritizes on-device training time over communication time in FL, as training time has a significantly greater impact  
(as in Tables \ref{training}-\ref{tb:end2end time}).

\textbf{Selection Process.} When a device joins the system, it notifies the server of its resource capabilities. Based on the resource capabilities of a specific device $u$, the server can compute the on-device training time and the utility drop using its public data $D_S$, i.e., using a backup device having the same device type and system resources with $u$.
Our selection process has two steps. First, the server will select all sub-networks with on-device training time less than the threshold $\mu_u$. 
Second, the server applies a brute-force search to find the optimal combination of a sub-network $\theta_u$ among the selected ones in the first step and $q_u \in [1, q^{max}_u]$, which minimizes the utility drop in Eq. \ref{selectionObjective}. The complexity of this brute-force search selection process is $\mathcal{O}(m \times q)$, where $m$ is the number of sub-networks and $q$ is the number of quantized bits.  This search is efficient and effective due to the small search space, allowing for early termination of the search when the utility drops below an acceptable level. 
Consequently, the server can identify the optimal sub-network $\theta_u$ and the number of quantized bits $q_u$ while ensuring efficient on-device training for the device $u$ with the least utility drop.


Notably, optimizing sub-networks is costly and would increase on-device computation and communication during federated training. While joint optimization may yield optimal sub-networks, our approach separates the tasks to reduce costs while maintaining high model utility with limited resources.

\subsection{FedX Training Protocol} 

\begin{algorithm}[t!]
\small
\caption{FedX Training Algorithm}\label{alg: FedX}
\begin{algorithmic}[1]
\STATE \textbf{Inputs}:    $N$ devices,    $T$ rounds,  quantization function $\mathbb{Q}$,  server's data $D_S$,  learning rates $\lambda$ (global) and $\eta$ (local),   $E$ epochs, quantization levels $\{q_u\}_{u \in [1, N]}$, devices' local data $\{D_u\}_{u \in [1, N]}$ 
\STATE \textbf{Outputs}: Global model $\theta$ 
\STATE \textbf{At the \text{Server}}:
\begin{ALC@g}
\STATE Initialize global model parameters $\theta$
\STATE One-time train the global model parameters $\theta$ with publicly available data at the server  $D_S$
\STATE Decompose and assign a sub-network  $ \{\theta^0_u \}_{u=1}^N$  to every device $u$ ($\theta^0_u \subset \theta$)  \textit{\cmt{Adaptive Model Decomposition in Fig.~\ref{fig:Training Protocol}}  }
\FOR{$t = 1, \ldots, T$}
\STATE Randomly select a set of devices $M$   
 \FOR{each device $u$ in $M$ \textit{\cmt{in parallel}} }
 \STATE Send the quantized sub-network $\mathbb{Q}(\theta_u^{t}, q_u)$ to device $u$ \textit{\cmt{Quantization $\mathbb{Q}$ in Fig.~\ref{fig:Training Protocol}}}
  \STATE $\theta_u^{t+1} =\textbf{LocalTraining}(u, \mathbb{Q}(\theta_u^{t}, q_u))$ \textit{\cmt{On-device Training in Fig.~\ref{fig:Training Protocol}}}
 \ENDFOR
 \STATE $\theta \leftarrow \mathcal{G} (\{ \theta_u^{t+1} \}_{u\in M})$ \textit{\cmt{Model Aggregation in Fig.~\ref{fig:Training Protocol}} }
 \STATE Fine-tune the global model:  $\theta^{t+1} \leftarrow \theta  - \lambda  \bigtriangledown_{\theta}  \mathcal{L} (\theta, D_S)$ \textit{\cmt{Model Fine-tuning in Fig.~\ref{fig:Training Protocol}}}
\ENDFOR
\end{ALC@g}
\STATE \textbf{LocalTraining}$(u, \theta_u^{t})$:
\begin{ALC@g}
\STATE Initialize: $\theta_i^0 \leftarrow \theta_u^{t}$%
\FOR{$e=1,2,\ldots, E$}
    \STATE $\theta_u^{e} = \theta_u^{e-1} - \eta \bigtriangledown_{\theta_u} \mathcal{L} (\theta_u^e, D_u)$ %
\ENDFOR
\STATE Return  $ \theta_u^E$   
\end{ALC@g}
\end{algorithmic} 
\end{algorithm}
\setlength{\textfloatsep}{10pt}

\begin{figure}[t]
      \centering
    \includegraphics[scale=0.031]{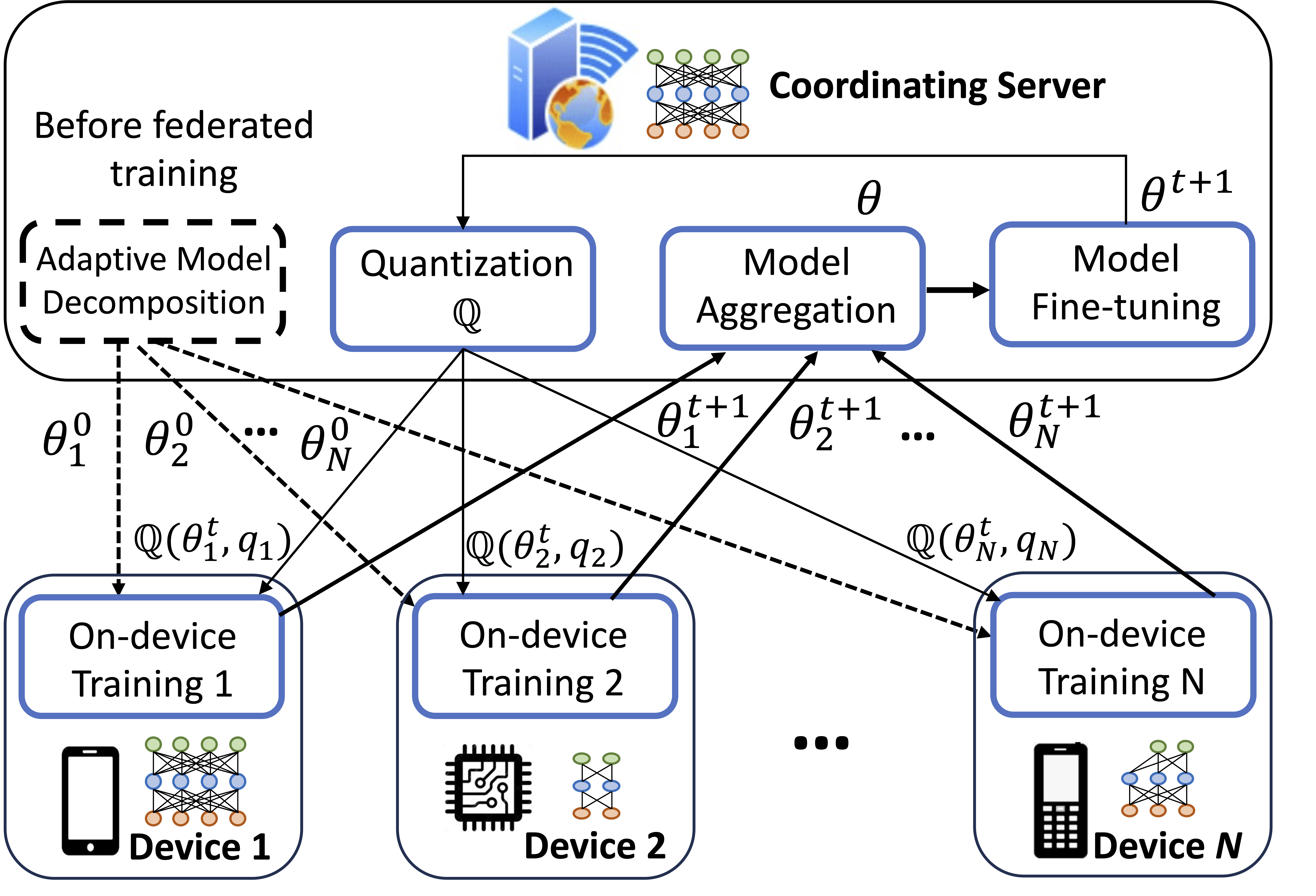} 
      \caption{FedX Training Protocol.}
      \label{fig:Training Protocol}
\end{figure}

 Fig.~\ref{fig:Training Protocol} and Alg.~\ref{alg: FedX}   present the protocol of FedX training to optimize the objective function in Eq. \ref{objectivefunction} and its pseudo-code. 
Initially, the server trains the global model $\theta$ on  public data $D_S$,  decomposes 
$\theta$ into nested sub-networks $\theta_u$, and then assigns them to devices $u$ based on computational power and processor bit-width. This assignment is performed once.



At each iteration $t$, the server selects a set of $M$ devices and sends different quantized sub-networks $\mathbb{Q}(\theta^t_u, q_u)$ with the numbers of quantized bits $\{q_u\}_{u \in M}$ to each device $u$ in $M$. The device $u$ uses the quantized sub-network as the local model $\theta_u = \mathbb{Q}(\theta^t_u, q_u)$.  Devices independently train and update the local sub-networks $\theta_u^t$ using their local data $D_u$ by minimizing the following loss function:
\begin{equation}
\small
    \theta_u^{t+1} = \arg \min_{\theta_u} \mathcal{L}(\theta_u, D_u),
    \label{localData}
\end{equation}
where $\theta_u = \mathbb{Q}(\theta^t_u, q_u)$.
Next, device $u$ sends the updated sub-network $\theta_u^{t+1}$ back to the server. 

After receiving all the updated sub-networks from the devices $u \in M$, the server aggregates them; that is, multiple sub-networks are updated together into the global model $\theta$. Although other aggregations can be used, we use FedAvg \cite{kairouz2021advances} as the aggregation function $\mathcal{G}$ without loss of generality:
\begin{equation}
\small
    \theta = \mathcal{G}( \{ \theta_u^{t+1} \}_{u \in M}) = (\sum_{u\in M} \theta_u^{t+1})/ \mathbb{H}_M,
    \label{Aggregation}
\end{equation}
in which $\mathbb{H}_M$ is a mask with the same size of the global model $\theta$. In $\mathbb{H}_M$, each value corresponds to the number of devices updating that particular parameter during iteration $t+1$. For example, in Fig.~\ref{fig:subnets-b}, all parameters in the sub-network $\theta_1$ (i.e., the gray area) are  updated  by device 1 tied to 
$\theta_1$ and device 2 tied to $\theta_2$. Therefore, the aggregated value of these parameters is $\frac{\theta_1 + \theta_2}{2}$. The additional parameters in $\theta_2$, which are in $\theta_1$ (i.e., the yellow part), are only updated by $\theta_2$ tied to device 2. Accordingly, their aggregated values are $\theta_2$.

Then, the server fine-tunes the aggregated global model with its public data $D_S$ by minimizing the loss function:
\begin{equation}
\small
    \theta^{t+1} = \arg \min_{\theta} \mathcal{L}(\theta, D_S) + \gamma \|\theta - \mathcal{G}( \{ \theta_u^{t+1} \}_{u \in M})\|_2
    \label{eq:regularization}
\end{equation}
where $\gamma$ is a regularization hyper-parameter.

By iteratively minimizing the loss in the devices' local data $\{D_u\}_{u \in [1, N]}$ in Eq. \ref{localData} and the loss in the server's publicly available data $D_S$ in Eq. \ref{eq:regularization}, FedX optimizes model utility on both the local and the public datasets. 

\textbf{Regularization as Knowledge Distillation.} Adding the regularization term $\|\theta - \mathcal{G}( \{ \theta_u^{t+1} \}_{u \in M})\|_2$ into the objective function in Eq.~\ref{eq:regularization} prevents significant shifts between local models trained on devices and the quantized sub-networks optimized using the server's data. Working as knowledge distillation, this regularization smoothens knowledge sharing across the server and devices in the model update aggregation (Eq. \ref{Aggregation}), further improving model utility. 

After fine-tuning, the server uses the updated global model $\theta^{t+1}$ in the next training round.



\subsection{FedX Inference}

During inference, each device uses the assigned sub-network that is 1) well-trained on data from the devices and the server and 2) suitable for its computational power and quantization level. In practice, we round up the number of quantized bits to Int8,  Float16, and a full precision (Float32) for storing tensors and performing computations on IoT devices.

\subsection{Convergence Analysis}

We analyze the convergence rate of FedX when optimizing a strongly convex and Lipschtiz continuous loss function $\mathcal{L}(\cdot, \cdot)$ to provide guidelines for practitioners to employ FedX in real-world applications. Our analysis indicates that under the optimization of FedX with a specific learning rate decaying method, the global loss $\mathcal{L}(\theta, D_S)$  will converge with the rate of $\mathcal{O}(\log T/T)$. Moreover, each device receives a sub-network from the converged global model $\theta^T$ at the inference time. Therefore, local quantized models will also converge near the global model with a marginal error in the quantization. To derive the analysis, we assume the following:

\begin{assumption}
    \label{assmpt:lipschitz}
    $\mathcal{L}(\cdot, \cdot)$ is $G$-Lipschitz function with respect to $\theta$, i.e., there exists a $G\in\mathbb{R}$ satisfying $\|\mathcal{L}(\theta, \cdot) - \mathcal{L}(\theta', \cdot)\|_2 \le G\|\theta - \theta'\|, \forall \theta, \theta'$.
\end{assumption}

\begin{assumption}\label{assmpt:convex}
    $\mathcal{L}(\cdot, \cdot)$ is a convex function, i.e., $\forall \theta, \theta', w\in[0, 1], \mathcal{L}\Big(w\theta + (1-w)\theta', \cdot\Big) \le w\mathcal{L}(\theta, \cdot) + (1-w)\mathcal{L}(\theta', \cdot)$.
\end{assumption}
\noindent These assumptions are typical in providing convergence analysis for FL algorithms in the previous works \cite{li2020convergence,wu2023faster}.

Consider the step $t \in [1, T\times E]$ and denote a device update $v_u^{t+1} = \theta_u^{t+1} - \eta\nabla_{\theta_u^{t}}\mathcal{L}(\theta_u^t, D_u)$. At step $t: t \% E = 0$, the update of the server is as~follows:
\begin{equation}
    \theta^{t} = \theta^{t-1} - \lambda\nabla_{\theta^{t-1}}\mathcal{L}(\theta^{t-1}, D_S) - \lambda\gamma[\theta^{t-1}- \mathcal{G}(v_u^{t})], \nonumber
\end{equation}
\noindent and $\theta^{t} = \theta^{t-1}$ at step $t: t \% E \neq 0$. Therefore, the server's updating process is a stochastic gradient descent process of a Lipschitz, convex function $\mathcal{L}(\cdot, \cdot)$. As a result, we can leverage Theorem 2 from Shamir et al. \cite{shamir2013stochastic} to derive the convergence rate of FedX for the global loss, as follows:

\begin{customthm}{1}\label{theo:conv_rate-app}
   For a convex function $\mathcal{L}(\theta)$, let $\theta \in \Theta$ such that $\|\Theta\|_2 \le Q$, $\theta^* = \arg\min_{\theta}\mathcal{L}(\theta)$, and $\theta^0$ is an arbitrary point in $\Theta$. Consider a stochastic gradient descent with the learning rate $\lambda_t = \frac{Q}{G\sqrt{t}}$. Then for any $T > 1$, the following is true
    \begin{align}
        \mathbb{E}(\mathcal{L}( \theta^T, D)) - \mathcal{L}( \theta^*,D) \le \mathcal{O}(\frac{QG\log(T)}{\sqrt{T}}).
    \end{align}
\end{customthm}

Hence, FedX ensures global model convergence at a rate of  $\mathcal{O}(\log(T)/\sqrt{T})$. While local model convergence is hard to quantify due to sub-network decomposition intractability, empirical results show consistent convergence across datasets, highlighting FedX's optimization efficiency.

\section{Evaluation}
\label{sec:Experiments}


We conduct extensive experiments to explore \textbf{1)} the interplay among quantized bits, model architectures, and utility, \textbf{2)} the impact of FedX's innovative components (i.e., nested sub-networks, quantization selection, and knowledge sharing among the server and devices) on utility, and \textbf{3)} how FedX saves resources on devices (e.g., CPU and memory), and reduces training, inference, and communication latency.


\begin{figure*}[t]
      \centering
       \includegraphics[scale=0.23]{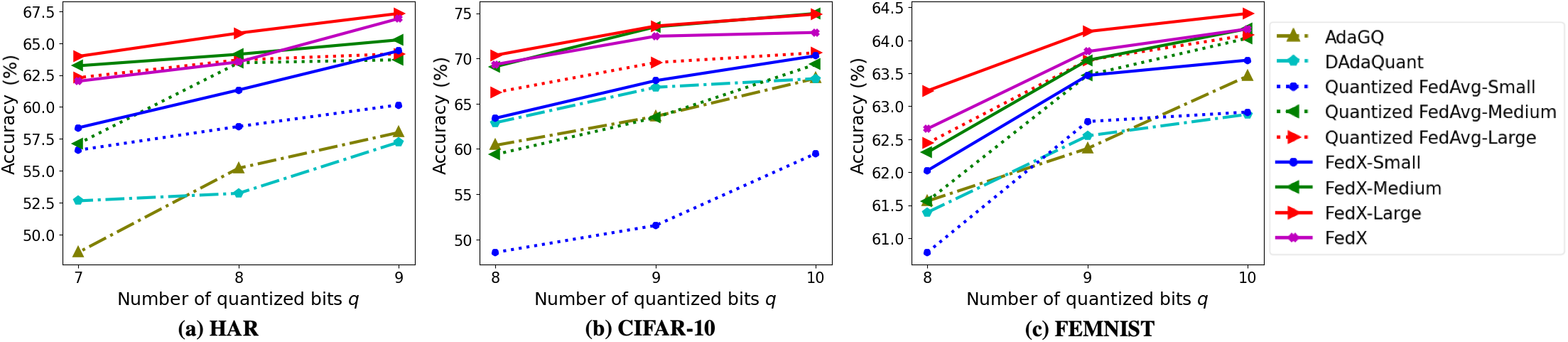} 
        \vspace{-5pt}
      \caption{Model accuracy on the HAR, CIFAR-10, and FEMNIST datasets.}
      \label{fig:results-3exps}
      \vspace{-5pt}
\end{figure*}
\setlength{\textfloatsep}{0pt}

\subsection{Devices}
We test FedX via simulations and an on-device prototype. For simulation, we use a GPU station with 4 NVIDIA GeForce Titan Xp GPUs (12G each), Intel Xeon E5-2637 v4 CPU (3.50GHz), Linux (Ubuntu 16.04), and CUDA 11.3. For on-device prototype, we employ three diverse IoT devices with heterogeneous resources, including \textit{1) Qualcomm QCS605 AI Kit,} which is equipped with Snapdragon™ QCS605 64-bit ARM v8-compliant octa-core CPU up to 2.5 GHz, Adreno 615 GPU, 8G RAM, and 16 GB eMMC 5.1 onboard storage; \textit{2) Nvidia Jetson Nano,} which features a quad-core ARM Cortex-A57 CPU with a clock speed of 1.43 GHz and 4 GB RAM; and \textit{3) Raspberry Pi 5,} which features a quad-core ARM Cortex-A72 CPU with clock speeds of 2.4GHz, and 4GB RAM.

\subsection{Datasets}

We conduct experiments on \textit{image classification} using Tiny-ImageNet \cite{le2015tiny}, CIFAR-10 \cite{krizhevsky2009learning}, and FEMNIST \cite{caldas2018leaf}, and on \textit{human activity recognition (HAR)} using the HAR dataset \cite{jiang2022flsys}. Tiny-ImageNet, CIFAR-10, FEMNIST, and HAR datasets have $100,000$, $50,000$, $600,000$, and $165,740$ training samples, respectively, with corresponding test sets of $10,000$, $10,000$,  $150,000$, and $28,770$  and class counts of $200$, $10$, $62$, and $5$. We simulate $100$, $100$, $1,000$, and $655$ devices, respectively. Devices and the server are allocated disjoint training classes with \textit{1)} $0$–$149$ (Tiny-ImageNet), \textit{2)} $0$–$7$ (CIFAR-10), \textit{3)} $0$–$45$ (FEMNIST), and \textit{4)} $0$–$3$ (HAR) on devices, and remaining classes at the server. Data is independent and identically distributed (IID) among clients. We also test FedX under non-IID settings with the HAR dataset.  Testing includes all classes, highlighting the effectiveness of knowledge sharing in FedX.

\subsection{Model Architectures}

We use MobileNetV3 \cite{howard2019searching} for image datasets and HAR-Wild \cite{jiang2022flsys} for the HAR dataset as global models. The MobileNetV3 model has five CNN blocks with varying layers ($d \in {2,3,4}$) and channels ($w \in {3,4,6}$), generating 243 nested sub-networks. To effectively demonstrate the impact stemming from various sub-networks on  utility and computation, we examine three sub-networks: \textbf{1)} \textit{small} ($d=2$, $w=3$, 2.41M parameters), \textbf{2)} \textit{medium} ($d=3$, $w=4$, 3.51M parameters), and \textbf{3)} \textit{large} ($d=4$, $w=6$, 6.39M parameters). The HAR-Wild model uses three 1D CNNs with batch norm and five fully connected layers, tailored for mobile devices with low complexity and memory.  We create sub-networks by reducing channels from 128 (large) to 64 (medium) and 32 (small), corresponding to 316K, 153K, and 76K parameters, respectively.




 \begin{figure}[t]
      \centering
       \includegraphics[scale=0.35]{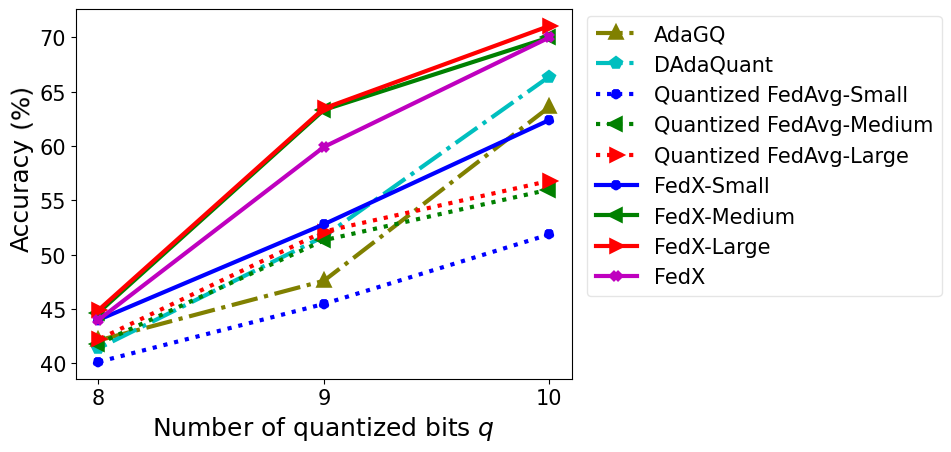} \vspace{-5pt} 
      \caption{Model accuracy  in  the Tiny-ImageNet dataset.} 
      \label{fig:results-tinyimagenet}
\end{figure}
\setlength{\textfloatsep}{5pt}

To reduce training overhead, we initialize MobileNetV3 pre-trained on ImageNet and HAR-Wild pre-trained on HAR-UCI \cite{anguita2012human}. FedX training uses a batch size of 32, a learning rate of $0.01$ with SGD optimizer, and a regularization rate $\gamma$ of $1e^{-4}$. 

\subsection{Baselines}

We evaluate FedX against baselines,  i.e., the  widely used FedAvg and state-of-the-art quantization-based FL methods for heterogeneous devices, including DAdaQuant \cite{honig2022dadaquant} and FL AdaGQ \cite{liu2023communication}. In addition, we compare variants of FedAvg and FedX with different sub-networks and server data availability, including 1) \textit{Quantized FedAvg-Small}: No server data or fine-tuning and  all devices use the same small sub-network; 2) \textit{Quantized FedAvg-Medium}: same as 1) but with   medium sub-networks;  3)  \textit{Quantized FedAvg-Large}: same as  1) but with  large sub-networks; 4) \textit{FedX-Small}: Server fine-tunes with data and regularization. All devices use the same small sub-network; 5) \textit{FedX-Medium}: same as   4) but with    medium sub-networks; 6) \textit{FedX-Large}: same as  4) but with   large sub-networks. Quantization is done at the server in these variants.

In \textit{FedX}, the server has data, fine-tunes  with regularization, and performs quantization. To be fair, 
we evaluate FedX, FedAvg, and  variants with DAdaQuant and AdaGQ, where the server has data, fine-tunes without regularization, and quantization occurs on devices as in their conventional settings.


\textbf{Model Accuracy.}
To evaluate model utility, we compute the average  accuracy across devices using their testing data, 
as follows: $Acc = \frac{1}{N} \sum_{u \in [1,N]} \big[ \frac{1}{|D_{u}^{test}|} \sum_{x\in D_{u}^{test}}  \mathbb{I}\big(
    \theta_u( x),y \big) \big]$,
where $\mathbb{I}\big(\theta_u( x),y \big)=1$ if $\theta_u( x)=y$ and $0$ otherwise. 
 $D_{u}^{test}$ consists of 
testing samples and its size $|D_{u}^{test}|$ at the device $u$.

\begin{figure}[t]
      \centering
       \includegraphics[scale=0.35]{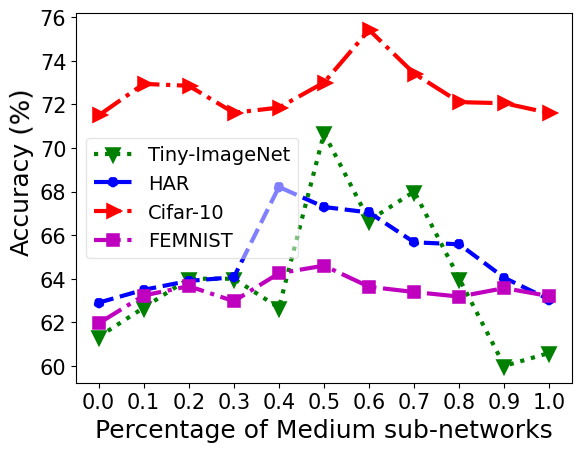} 
        \vspace{-5pt}
      \caption{Varying percentage of medium and small sub-networks.}
      \label{fig:results-percentage}
\end{figure}
\setlength{\textfloatsep}{5pt} 


\subsection{Model Utility Evaluation}  

Fig.~\ref{fig:results-tinyimagenet} illustrates model utility on the Tiny-ImageNet dataset across numbers of quantized bits, and different FedX and baseline versions. Overall, FedX-Medium and FedX-Large perform the best.
FedX achieves a better balance between sub-network sizes and the numbers of quantized bits. Large sub-networks do not consistently lead to a notable accuracy improvement. 
There is a marginal difference in model accuracy between FedX-Medium and FedX-Large across  numbers of quantized bits. However, using a sub-network that is too small would significantly damage the model utility, i.e., FedX-Small. This illustrates a non-trivial trade-off between sub-network sizes and the number of quantized bits in FL. FedX mitigates this problem by identifying the  most suitable sub-network and  quantized bits for each device, minimizing performance degradation  
while adhering to the device's resource constraints.

Given the marginal accuracy gap between FedX-Medium and FedX-Large, but a significant gap with FedX-Small, for the following experiments, we use only the \textit{medium and small sub-networks}  in FedX. In this setup, half of the devices use the same small sub-network and half uses the medium one.  
We change this proportion in our ablation study to understand the trade-off between nested sub-networks and quantization selection. DAdaQuant and AdaGQ also use two sub-networks for a fair comparison. As shown in Fig.~\ref{fig:results-tinyimagenet}, FedX outperforms DAdaQuant and AdaGQ in terms of model utility, with average improvements of 2.28-3.01\%, 8.20-12.28\%, and 1.9-4.76\%,
when the number of quantized bits increases from 8 to 10. 
Also, 
FedX outperforms the most accurate Quantized FedAvg-Large by 2.19\%, 2.49\%, and 11.56\% with 8, 9, and 10 quantized bits, respectively. This gain stems from nested sub-networks and an efficient training algorithm that boosts knowledge transfer and reduces device quantization costs, enhancing performance.

We observe similar trends   with HAR, CIFAR-10, and FEMNIST datasets (Fig.~\ref{fig:results-3exps}), where FedX-Medium and FedX-Large  outperform the baselines, demonstrating FedX's generalization across datasets and models. These results are from real-world data, considering heterogeneous quantization, limited resources, and data distribution differences, with only a $3.6\%$ drop compared to homogeneous devices without constraints \cite{jiang2022flsys}.

\begin{table}[t] 
\centering 
\caption{FedX Training on Qualcomm AI Kit.} \vspace{-5pt}
\resizebox{0.44\textwidth}{!}{
\begin{tabular}{|c|c|c|c|c|c|}
 \hline
 \shortstack{Dataset}  &  \shortstack{Model\\Size}  &  \shortstack{CPU\\(\%)} & \shortstack{RAM\\(MB)}  & \shortstack{Training \\time (s)}  & \shortstack{Communication \\time (s)}  \\ 
 \hline
\multirow{3}{*}{Tiny-ImageNet} & Small &	41.4&	922.4&	34.9&	2.29\\ 
                    &Medium &	44.8&	2005.7&	57.3&	2.96\\ 
                    &Large&	45.5&	2139.9&	77.5&	3.44\\ 
                    \hline
\multirow{3}{*}{CIFAR10} & Small &	35.3	&1072.4	&24.3&	2.05\\ 
                    &Medium &	36.5	&1197.5	&38.0&	2.69\\ 
                    &Large&	36.8	&1322.7	&49.8&	3.17\\ 
                    \hline
\multirow{3}{*}{FEMNIST} & Small &	35.8&	478.4&	24.3&	2.13\\ 
                    &Medium &	36.4&	761.9&	37.5&	2.77\\ 
                    &Large&	36.9&	920.8&	49.9&	3.25\\             
 \hline
\multirow{3}{*}{HAR} & Small &	31.6&	232.9&	2.6&	0.01\\ 
                    &Medium &	47.2&	245.4&	2.8&	0.04\\ 
                    &Large&	52.8&	252.1&	4.4&	0.13\\   
                     \hline
\end{tabular}}
\label{training}  
\end{table}
\setlength{\textfloatsep}{5pt}





\subsection{Ablation Study} 

To comprehensively examine the effects of FedX's innovative components, we conduct ablation studies on 1) the trade-off between nested sub-networks and quantization selection by varying the percentages of different types of sub-networks and  quantization levels, 2) the effects of knowledge-sharing among the server and devices, with and without regularization.

\textbf{Nested Sub-networks and Quantization Selection.} 
We vary the percentage of clients with medium sub-networks in FedX from $0\%$ to $100\%$, using smaller quantization levels (i.e., 9 for CIFAR-10, FEMNIST, Tiny-ImageNet; 8 for HAR) for medium sub-networks and larger levels (11 and 10, respectively) for small sub-networks. Given different percentages and datasets,  model utility typically peaks when  $40-60\%$ of medium sub-network clients are used, achieving $70.67\%$, $68.21\%$, $75.41\%$, and $64.60\%$ on Tiny-ImageNet, HAR, CIFAR-10, and FEMNIST (Fig.~\ref{fig:results-percentage}). 
Model utility drops by $2.05\%-5.31\%$ at higher percentages (i.e., more clients) of medium sub-network clients with smaller quantization levels.
These observations highlight the need to balance sub-network sizes and quantization for optimal performance and memory efficiency.

\begin{figure}[t]
      \centering
       \includegraphics[scale=0.39]{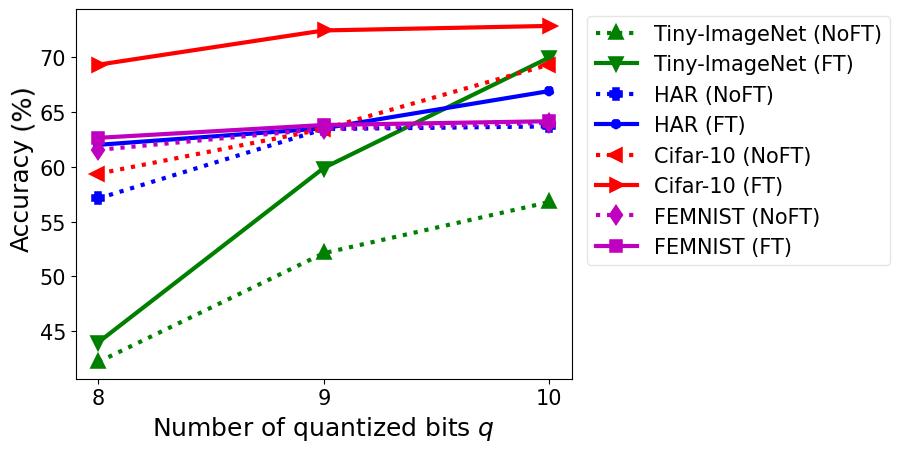} \vspace{-5pt}  
      \caption{FedX with and without Fine-tuning (FT and NoFT).} 
      \label{fig:results-finetune}
\end{figure}

\textbf{Knowledge Sharing via Regularization.} We compare models with and without fine-tuning, reflecting the presence or absence of knowledge sharing among the server and devices. Fig.~\ref{fig:results-finetune} shows consistently higher model utility with server fine-tuning. This is because when the server and the devices have different class distributions, the fine-tuning process fosters knowledge sharing among them, boosting FedX performance.



 \textbf{Non-IID Settings.} We conduct  experiments using the HAR dataset, where the number of samples per class per device follows an $\alpha$-symmetrical Dirichlet distribution and $\alpha \in [0.01, 1, 100]$.  FedX achieves $79.96\%$ accuracy when the data is extremely diverse (i.e., the concentration parameter $\alpha=0.01$), compared with $68.21\%$ in the IID setting. This utility improvement is because each device has fewer classes resulting in better local minima for their models. We observe a similar trend with different levels of non-IID, i.e., $\alpha \in [1, 100]$.

\begin{table}[t] 
\centering 
\caption{FedX Training on Raspberry Pi 5.} \vspace{-5pt}
\resizebox{.44\textwidth}{!}{
\begin{tabular}{|c|c|c|c|c|c|}
 \hline
 \shortstack{Dataset}  &  \shortstack{Model\\Size}  &  \shortstack{CPU\\(\%)} & \shortstack{RAM\\(MB)}  & \shortstack{Training \\time (s)}  & \shortstack{Communication \\time (s)}  \\ 
 \hline
\multirow{3}{*}{Tiny-ImageNet} & Small &	93.6&	479.5&	17.6&	2.29\\ 
                    &Medium &	92.6&	576.8&	32.1&	2.96\\ 
                    &Large&	91.9&	696.2&	43.4&	3.44\\ 
                    \hline
\multirow{3}{*}{CIFAR10} & Small &	89.3	&638.9	&7.5&	2.05\\ 
                    &Medium &	90.6	&632.2	&11.6&	2.69\\ 
                    &Large&	90.6	&632.9	&14.5&	3.17\\ 
                    \hline
\multirow{3}{*}{FEMNIST} & Small &	87.3&	368.5&	7.5&	2.13\\ 
                    &Medium &	88.8&	418.6&	11.1&	2.77\\ 
                    &Large&	89.9&	418.7&	13.7&	3.25\\             
 \hline
\multirow{3}{*}{HAR} & Small &	79.8&	279.0&	0.4&	0.01\\ 
                    &Medium &	84.8&	287.0&	1.2&	0.04\\ 
                    &Large&	83.4&	296.8&	2.7&	0.13\\   
                     \hline
\end{tabular}}
\label{rpi}  
\end{table}
\setlength{\textfloatsep}{5pt}

\setlength{\textfloatsep}{5pt}

%

\begin{table*}[h] 
\centering 
\caption{FedX Training on Nvidia Jetson Nano.} \vspace{-5pt}
\resizebox{0.72\textwidth}{!}{
\begin{tabular}{|c|c|c|c|c|c|c|c|c|c|c|c|c|c|}
 \hline
 \multirow{2}{*}{Dataset}  &  \multirow{2}{*}{\shortstack{Model\\Size}}  &  \multicolumn{4}{c|}{CPU Training}    &  \multicolumn{6}{c|}{GPU Training}    & \multirow{2}{*}{\shortstack{Communication \\time (s)}}  \\ 
\cline{3-12} 
 &    &  \shortstack{CPU\\(\%)} & \shortstack{CPU\\RAM\\(MB)}  & \shortstack{CPU\\Power\\(MW)} & \shortstack{Training \\time (s)}
 &  \shortstack{GPU\\(\%)} & \shortstack{GPU\\RAM\\(MB)}  & \shortstack{GPU\\Power\\(MW)}  & \shortstack{CPU\\RAM\\(MB)}  & \shortstack{CPU\\Power\\(MW)}  & \shortstack{Training \\time (s)} &  \\ 
 \hline
\multirow{3}{*}{Tiny-ImageNet} & Small &92.6&	98.4&	2267&	122.4&  99.7&   579&1275&616&687&	4.6&		2.29\\ 
                                &Medium &93.6&	129&	2375&	216.3&	99.7&	687&1338&611&663&	6.9&		2.96\\ 
                                &Large&97.9&	151&	2370&	241.6&	99.6&	755&1450&530&885&	16.9&		3.44\\ 
                    \hline
\multirow{3}{*}{CIFAR10} & Small &90.1&	141&	1914&87.4&	99.7&	511&722	&597&788&	5.6&		2.05\\ 
                        &Medium& 90.8&	144&	1781&189.1&	99.7&	549&1057	&604&900&	8.5&	2.69\\ 
                         &Large&82.3&	144&	1800&241.3&	99.7&573&	943	&583&909&	20.2&		3.17\\ 
                    \hline
\multirow{3}{*}{FEMNIST} & Small &92.4&	72.9&	1933&90.6&	99.7&	509&	1071&629&850&	5.3&		2.13\\ 
                    &Medium &92.8&	79.5&	2014&	173.9&	99.0&	527&886&628&851&	8.0&		2.77\\ 
                    &Large&92.1&86.2&	2020&	233.6&	99.7&	561&1006&616&917&	18.9&	3.25\\ 
                      \hline
\multirow{3}{*}{HAR} & Small &88.8&	65.2&	1307&2.8&	52.1&	474&	287&665&828&	0.5&0.01\\ 
                    &Medium &92.7   &	67.2&	2371&	2.7&	85.5&	478&647&662&672&	0.9&		0.04\\ 
                    &Large&91.9&	73.0&	2364&	6.5&	99.7&	520&1450&660&783&	0.8&	0.13\\    

 \hline
\end{tabular}
}
\label{jetson}  

\end{table*}
\setlength{\textfloatsep}{0pt}

\begin{figure*}
 \begin{minipage}[b]{.62\linewidth}
    \centering
    \resizebox{.9999\textwidth}{!}{\begin{tabular}{|c|c|c|c|c|c|c|c|}
 \hline
Dataset &  Mechanism &  \shortstack{On-device \\computation \\time (s)}  & \shortstack{Communication\\time (s)}  &\shortstack{ Server\\ quantization\\time (s)}   & \shortstack{Aggregation\\time (s)} & \shortstack{Fine-\\tuning\\time (s)} & \shortstack{Total\\time\\ (s)}  \\ 
 \hline
\multirow{3}{*}{Tiny-ImageNet} & DAdaQuant & 112.4 & 1.23 & N/A & 0.032 &  5.07 &118.73 \\ 
                               & AdaGQ & 112.4 & 1.72 & N/A &  0.032 &  N/A  & 114.15\\
                                & \shortstack{Quantized FedAvg} & 77.5 &  3.44 & 14.52 & 0.032 & N/A & 95.5  \\ 
                                 & FedX &77.5 &   3.44 & 4.14 & 0.032  &  5.07 &90.2 \\ 
                    \hline
\multirow{3}{*}{CIFAR10} &  DAdaQuant& 74.3 & 1.13  &  N/A & 0.032& 4.11&79.57\\ 
                        &  AdaGQ& 74.3 & 1.59  & N/A & 0.032& N/A& 75.92\\ 
                         &\shortstack{Quantized FedAvg}  & 49.8 &  3.17& 9.58 & 0.032 & N/A&62.6\\ 
                        & FedX& 49.8 & 3.17 & 4.36 & 0.032 &4.11&61.5\\ 
                    \hline
\multirow{3}{*}{FEMNIST} & DAdaQuant& 75.0 & 1.16 & N/A &0.033 & 3.73 &79.92 \\ 
                         & AdaGQ& 75.0 & 1.63 & N/A &0.033 & N/A & 76.66\\ 
                         &\shortstack{Quantized FedAvg }& 49.9 & 3.25 & 8.05  & 0.033 & N/A&61.2\\ 
                        & FedX & 49.9 & 3.25 & 3.63 &0.033 & 3.73  &60.5\\                     
 \hline
 \multirow{3}{*}{HAR} & DAdaQuant& 5.7 & 0.05  & N/A & 0.031& 0.47 & 6.25\\ 
                      & AdaGQ& 5.7 & 0.07 & N/A & 0.031 & N/A & 5.80 \\ 
                         &\shortstack{Quantized FedAvg }& 4.4 &0.13  & 4.11  & 0.031 & N/A& 8.67\\ 
                        & FedX & 4.4 & 0.13 & 2.78 &0.031 &  0.47 & 7.81\\                     
 \hline
\end{tabular}}
    \captionof{table}{End-to-end Operation Time.}
    \label{tb:end2end time} 
  \end{minipage}\hfill
  \begin{minipage}[b]{.41\linewidth}
    \centering
    \includegraphics[scale=0.4]{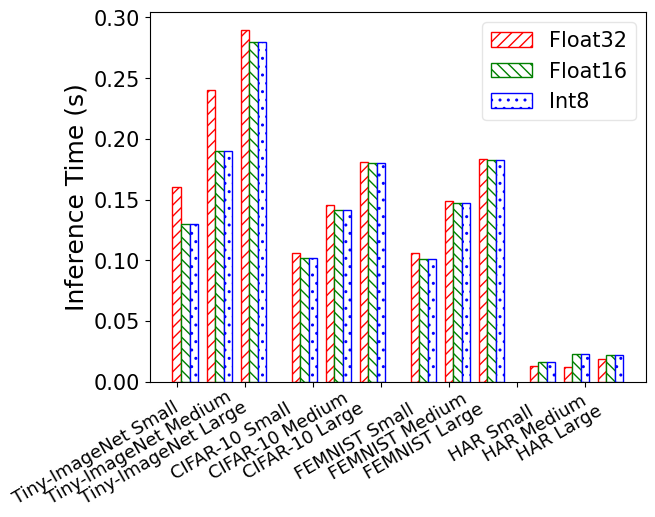} \vspace{-7.5pt}
    \captionof{figure}{FedX's On-device Inference time.}
     \label{infer}
  \end{minipage}
\end{figure*}
\setlength{\textfloatsep}{0pt}

\subsection{System Performance}

\textbf{On-device Training.}
\x{The experiments are conducted over  small, medium, and large sub-networks across   datasets.} We train on $750$ samples with a batch size of $32$ (i.e., one   training round) and report the mean of $20$ measurements in each dataset. 

Tables \ref{training}, \ref{rpi}, and ~\ref{jetson} show the FedX resource consumption, training, and communication times across devices and datasets.
Training times range from 0.5 to 77.5 seconds (s), while communication times (i.e., the round-trip time of sending and receiving sub-networks with the network bandwidth of 30 mbps) span  0.01s to 3.44s.
These training and communication times are fast and practical.
In addition,  CPU usage during training ranges from 31.6\% to 97.9\%, with maximum RAM usage from 65.2 MB to 2,139.9 MB, indicating on-device training feasibility.
Notably, GPU training on Nvidia Jetson Nano via CUDA uses comparable power as CPU, but it is $3$ to $31\times$  faster than CPU training, especially for larger datasets and models (i.e., 14 to 31 $\times$ for Tiny-ImageNet compared to 3 to 8 $\times$ for   HAR), highlighting FedX's efficiency with growing GPU adoption in IoT devices. This observation highlights FedX's efficiency with growing GPU adoption in IoT devices. 

\textbf{On-device Inference.} 
Since FedX supports Int8, Float16, and Float32, we report the inference time using the maximum supported  bits.  Fig.~\ref{infer} shows the times on the Qualcomm AI Kit across sub-networks and quantized bits. While the small HAR model shows negligible differences, other models benefit from Int8 and Float16, reducing latency. For Tiny-ImageNet, Int8 and Float16 small sub-networks take $0.13$s compared to $0.16$s for Float32. Medium and large Int8 sub-networks take $0.19$s and $0.27$s, with Float16 slightly higher. Overall, the inference time is under $0.3$s, showing FedX's efficiency and feasibility.



\textbf{End-to-end Operation Time.} 
FedX's end-to-end operation includes on-device computation, communication,  quantization, aggregation, and fine-tuning. Table~\ref{tb:end2end time} shows the average  operation time over 20 rounds. Since the devices run in parallel, on-device computation and communication are measured by the largest sub-network on Qualcomm AI Kit, which is the slowest among the IoT devices. For a fair comparison, we evaluate DAdaQuant, AdaGQ, and Quantized FedAvg with the large sub-network, as it offers the highest accuracy and these methods do not vary model sizes. 

DAdaQuant and AdaGQ have on-device computation overhead from on-device quantization, causing 
on-device computation times from 30\% to 49\% longer than in FedX and Quantized FedAvg-Large. 
In FedX, the quantization time is between $3.63$ and $2.78$s per round, making it $3.51\times$, $2.20\times$, $2.22\times$, and $1.48\times$  faster than Quantized FedAvg-Large in the Tiny-ImageNet, CIFAR-10, FEMNIST, and HAR datasets, respectively.
The server consumes a negligible aggregation time, lasting about $0.03$s. For fine-tuning, the time for DAdaQuant and FedX is measured by training the same sample size across datasets, while Quantized FedAvg-Large does not perform fine-tuning after quantization.

The total operation time  on the HAR dataset is negligible due to HAR-Wild model's small number of parameters.  
In other datasets,
FedX is faster than DAdaQuant by 22.7\% to 24.3\%, AdaGQ by 19.0\% to 21.1\%, and Quantized FedAvg-Large by 1.1\% to 5.5\%, even though Quantized FedAvg-Large does not  fine-tune. FedX works efficiently on-device, enhances end-to-end operation, and offers a more practical FL system.

\x{More importantly, FedX converges faster than the baselines due to enhanced  
knowledge sharing 
between the server and devices. For example, in the Tiny-ImageNet, FedX, FedAvg, DAdaQuant, and AdaGQ require $6$, $15$, $32$, and $30$ rounds to converge. 
This result is significant as FedX reduces training times per round and minimizes the  number of training rounds.}

\textbf{Remark.} FedX offers four key advantages: \textbf{(1)} FedX has higher model utility, with a lower resource consumption and quantization time. \textbf{(2)} FedX's quantization is $3.51\times$ to $8.43\times$ faster than FedAvg, DAdaQuant, and AdaGQ. \textbf{(3)} Thanks to knowledge sharing between the server and  devices, FedX converges quickly.  \textbf{(4)} With the on-device training time under $77.5$s,  on-device inference time under $0.3$s, and  communication time 
under $3.44$s, FedX is feasible in  real-world  applications.


\section{Related Work}
\label{app: Related Works}



Device heterogeneity poses a key challenge in FL, with solutions like dropping or partial updates leading to resource waste and utility drop \cite{chai2020tifl,zhang2023timelyfl}. Two main approaches aim to tackle this challenge   and improving communication efficiency.  


First, compression methods, such as quantization,  pruning, sparsification, and sampling \cite{shlezinger2020uveqfed,zhang2023preserving,Jiang_Borcea_2023,hamer2020fedboost,abdi2020quantized,vithana2023model}, reduce the amount of data transfers but may impact utility or increase computational overhead. In contrast, FedX  improves utility and reduces overhead with server-side quantization and knowledge transfer. 


Second, knowledge distillation (KD) transfers knowledge from large to smaller models without sacrificing utility \cite{li2019fedmd,zhang2023towards,zhu2021data,wu2022communication,chen2023metafed,xu2023feddk}. Two main approaches are 1) data-additional KD, combining public and private data for training, and 2) data-free KD, transmitting prediction scores without public data. In FedX, we employ quantization and knowledge transfer, to improve knowledge sharing while reducing overhead.


 Using FL in IoT devices poses   challenges in terms of communication efficiency and   heterogeneity. Proposed solutions include codistillation that exchanges model predictions to train local models, device-inclusion that assigns models based on resource capabilities, and device selection using resource-based criteria \cite{sodhani2020closer,cho2023communication,fu2023client}. However, they lack adaptive quantization to balance communication, computation, and utility.


\section{Conclusion}

This paper presented FedX, a novel FL system with adaptive model decomposition and quantization for heterogeneous IoT devices. FedX assigns smaller sub-networks with higher quantization to resource-limited devices and larger ones with lower quantization to resource-rich devices, enhancing knowledge sharing and resource efficiency.  Knowledge sharing occurs by aggregating nested sub-networks through the global model each round. Optimized and diverse sub-network sizes significantly reduce computational overhead compared to single large models. 
FedX ensures global model convergence. Overall, FedX  improves utility and system effectiveness while reducing training, quantization, and inference overhead, making it suitable for heterogeneous IoT environments.


\section*{Acknowledgments}
This research was supported in part by the National Science Foundation
(NSF) under Grants No. CNS 2237328, DGE 2043104, and CNS-1935928.

\bibliographystyle{IEEEtran}
\bibliography{IEEEfull_FedX}



\end{document}